\documentclass[11pt,letterpaper,logo,twocolumn]{style}

\usepackage[numbers]{natbib}
\usepackage{graphicx}
\usepackage{booktabs}
\usepackage{amsmath,amsfonts,amssymb}
\usepackage{subcaption}
\usepackage{multirow}
\usepackage{colortbl}
\usepackage{listings}
\usepackage{xparse}
\usepackage{fontawesome5}
\usepackage{threeparttable}

\graphicspath{{./}{fig/}{figures/}{plot/}{pdf/}{table/}}
\usepackage{amsthm}
\usepackage{tcolorbox}
\tcbuselibrary{skins,breakable}
\tcbuselibrary{listingsutf8}
\usepackage{titletoc}
\usepackage{pifont}
\usepackage{mathtools}
\usepackage{bbm}
\usepackage{makecell}
\usepackage{adjustbox}

\usepackage{xspace}
\newcommand{\heading}[1]{\vspace*{1mm}\noindent\textbf{#1}}
\newcommand{\benchname}{PhysTool-Bench\xspace}

\title{Beyond APIs: Probing the Limits of MLLMs in Physical Tool Use}
\runningtitle{Beyond APIs: Probing the Limits of MLLMs in Physical Tool Use}
\PublicDate{2026-06-09}

\author{%
  {\Authfont
    \textbf{Zhixin Ma}\textsuperscript{1}\equal \quad 
    \textbf{Yutong Zhou}\textsuperscript{2}\equal \quad
    \textbf{Yongqi Li}\textsuperscript{2}\advisor \quad
    \textbf{Chong-Wah Ngo}\textsuperscript{1} \quad
    \textbf{Wenjie Li}\textsuperscript{2} \quad
  }\\
  {\Affilfont
    \textsuperscript{1} Singapore Management University \quad
    \textsuperscript{2} The Hong Kong Polytechnic University \\
    \texttt{\{zhixinma97, yutongzhou714, liyongqi0\}@gmail.com}
  }
}

\begin{document}

\begin{abstract}
Multimodal Large Language Models (MLLMs) excel at utilizing digital APIs and increasingly serve as the ``brain'' of embodied AI, instructing robots to interact with the physical world. In such embodied settings, a central capability is the use of physical tools, which underpins MLLMs' ability to assist humans in real-world tasks. Despite the importance, MLLMs' proficiency in physical tool use remains largely unexplored. To address this gap, we introduce \textit{\benchname}, the first physical tool-use benchmark designed to evaluate MLLMs' ability to comprehend real-world scenarios, identify physical tools, and plan their use. \textit{\benchname} comprises 2,510 queries over 2,678 real-world physical tools spanning diverse domains, including manufacturing, electrical work, agriculture, and healthcare. Concretely, models are evaluated along two primary dimensions: 1) recognizing all physical tools present in the scene, and 2) planning the tool selection and use sequence based on the instruction and visual context. Across 13 leading MLLMs, even the strongest model (Gemini-3.1-Pro) identifies only 58.7\% of tools in a scene and completes merely 21.0\% of queries end-to-end. Our analysis reveals a two-level deficit: MLLMs struggle to perceive tools in realistic scenes, and the much larger drop at the planning stage further indicates a lack of functional commonsense for mapping perceived tools onto task semantics, pinpointing a critical bottleneck for the development of practical embodied AI.

\end{abstract}

\newcommand{\TitleLinks}{%
\centering
    \vspace{6pt}
    {\noindent\absfont\fontsize{11}{13}\selectfont
    \faGithub\ Project Page: \url{https://github.com/ModalityDance/PhysTool-Bench}\par}%
}



\maketitle

\section{Introduction}

\begin{figure}[h]
  \centering
  \includegraphics[width=1\linewidth]{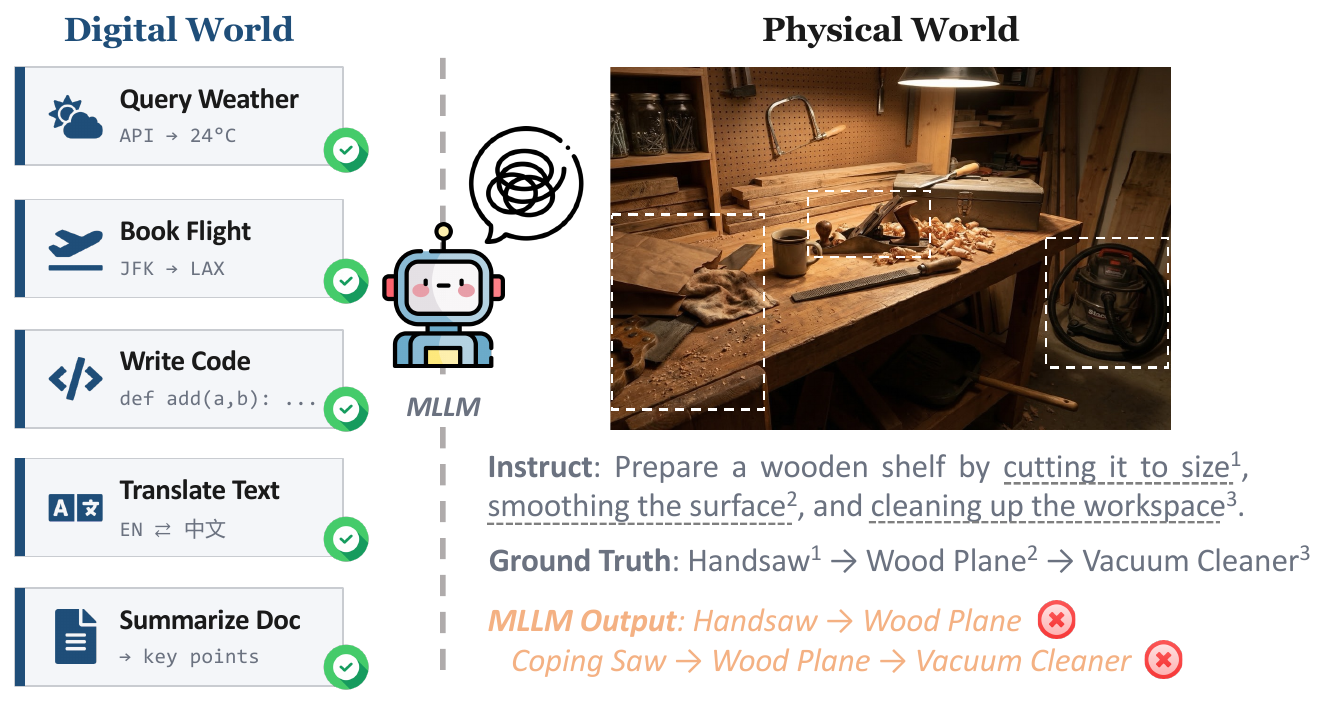}
  \caption{The capability divide between digital and physical tool use. MLLMs solve structured digital tasks reliably via APIs (left), but struggle with the visual reasoning and physical commonsense required to select and sequence tools in real-world scenes (right). \textit{PhysTool-Bench\xspace} evaluates exactly this physical-world capability. Target tools are highlighted only for illustration.}
  \label{fig:intro}
\end{figure}

The ability to use tools has long been a core capability of intelligence, and Large Language Models (LLMs) have recently shown remarkable progress along this dimension. State-of-the-art LLMs now function effectively as autonomous digital agents, using software APIs to book flights, query databases, and navigate the web~\cite{yun23apibench,qin24toolllm}. However, these successes are confined to the digital world with APIs. As an essential step toward deploying AI to assist human society, the capability of these models to follow instructions and utilize tools in the physical world must also be rigorously assessed.

Multimodal LLMs (MLLMs) are increasingly regarded as the reasoning core of embodied AI~\cite{li23manipllm}. By integrating visual perception with language comprehension, MLLMs empower embodied agents to ground high-level instructions, such as ``bring me the red mug on the kitchen counter'', into actions that robots can execute. Recent systems have shown strong performance on indoor navigation~\cite{danny23palme} and object manipulation~\cite{liu24robomamba}, and the benchmarks driving this progress have largely focused on the same two capabilities~\cite{xiang20sapien,mu21maniskill}. Yet tool use in the physical world, arguably the next frontier for embodied AI, has received far less attention. Specifically, how well current MLLMs can recognize, comprehend, and utilize physical tools remains an open question.

To answer this question, we introduce \textit{\benchname}, a benchmark dedicated to evaluating physical tool use. \textit{\benchname} contains 2,510 queries over 2,678 distinct physical tools, drawn from manufacturing, electrical work, agriculture, healthcare, and beyond. Each query pairs a natural-language instruction with an image of a realistic environment, such as a workshop or kitchen, where the model must identify the appropriate tools for the task. As illustrated in Figure ~\ref{fig:intro}, given an instruction such as ''prepare a wooden shelf...'', the model must select the correct tools (a handsaw, plane, and vacuum cleaner) in the right order, while rejecting visually or functionally similar alternatives. We evaluate MLLMs on two progressive tasks: Task I (Physical Tool Recognition) asks the model to enumerate every tool visible in the scene; Task II (Tool Selection and Action Planning) further requires it to select the necessary tools and place them in the correct execution order, given the instruction. Together, the two tasks disentangle what the model can see from what it can reason about.

\textit{\benchname} mirrors the visual and conceptual complexity of real-world environments. Each scene contains on average 8.6 tools, of which only 3.1 are required by the instruction; the remaining items are everyday tools that may be visually or functionally related to the targets. 86.9\% of queries further require multiple tools to be applied in a specific order, jointly evaluating selection and sequential planning. To capture both axes of difficulty, we report set-level F1 for tool selection and strict Exact Match (EM), which requires the predicted tools to match the ground-truth set \emph{and} execution order. The full dataset is curated through a multi-stage quality-control pipeline (§\ref{subsec:dataset_pipeline}), and a human reference study confirms its quality: on queries rated highly familiar by an annotator, human EM reaches \textbf{75\%}, indicating that the ground truth aligns with informed human judgment.

We benchmark 13 leading MLLMs on \textit{\benchname}, spanning commercial models (GPT-4o, GPT-5.2, Gemini-3.1-Pro, Qwen3-VL-Plus) and open-source models (Qwen3-VL, InternVL, Kimi-VL, DeepSeek-VL, and others). Four findings stand out.
(i) \textbf{Recognition is non-trivial.} Even the strongest model identifies only 58.7\% F1 of the tools in a scene; most open-source models miss more than half.
(ii) \textbf{Action planning is far harder.} Gemini-3.1-Pro succeeds on merely 21.0\% (EM) of queries, with EM collapsing from 34.5\% on two-tool queries to 0.5\% on queries requiring six or more.
(iii) \textbf{Functional confusion drives failures.} 42--61\% of errors stem from substituting target tools with functionally similar alternatives that are visible in the scene; a specialized open-vocabulary detector (Grounding DINO) even outperforms the best MLLM in recall by 13.4\,pp, indicating that the bottleneck is physical commonsense, not perception.
(iv) \textbf{The model gap is real.} Averaged across all familiarity levels (including unfamiliar domains), the human annotator reaches \textbf{38\% EM}, far exceeding the best MLLM (21.0\%), confirming the gap reflects model capability rather than task ambiguity.

In summary, our contributions are as follows:
\begin{itemize}
    \item \textbf{A new dimension for evaluating MLLMs.} We introduce \textit{\benchname}, the first benchmark dedicated to physical tool use. This capability bridges digital tool mastery and real-world embodied deployment, yet has remained largely unexamined despite recent progress in Embodied AI.
    
    \item \textbf{A diagnostic evaluation framework.} Our two-task design, separating recognition from instruction-conditioned selection and planning, isolates failures along the perception-to-reasoning pipeline. The benchmark provides verified ground truth across \textbf{2,510 queries} spanning \textbf{2,678 tools} in everyday domains from manufacturing to healthcare.
    
    \item \textbf{A pointed empirical diagnosis.} Across 13 state-of-the-art MLLMs, we find that the bottleneck in physical tool use is not raw perception but \textbf{functional commonsense}: even when models correctly perceive a scene, they fail to map tools onto task semantics. This points to physical commonsense as the central research direction for practical embodied AI.
\end{itemize}

\section{Related Work}

\subsection{Benchmarks for Digital Tool Learning}
Recent studies have demonstrated the power of LLMs to master the use of external tools to solve complex problems \cite{schick23toolformer, ReAct}. Early methods have confirmed the potential of tool learning in overcoming limitations of LLMs as a language processor while maintaining its generality \cite{schick23toolformer}. 

Encouraged by the promising future of tool learning, a variety of benchmark and evaluation studies have been established to systematically define the problem. General benchmarks typically evaluate LLMs' ability in tool selection and tool calling across various APIs and their diverse use cases \cite{patil24gorilla}. Subsequent studies expand the scope to include action planning and response generation stages \cite{qin24toolllm}, while later version has evolved to balance between stability and reality via a virtual API server \cite{guo24stabletoolbench}. 
However, these existing benchmarks are predominantly confined to textual modalities and digital API environments. They fail to assess how agents visually perceive real-world scenarios and manipulate physical tools.

\subsection{Evaluations for Embodied Action Planning}

The transition from digital assistants to physical robots necessitates the evaluations of how well high-level reasoning can be grounded in robotic affordances. . Since the advent of SayCan \cite{ahn22saycan} which introduced pre-trained robotic value functions to assess the feasibility of each planned step, researchers have been working on bridging the gap between LLM’s high-level semantic knowledge and long-horizon task planning and completion in real world. While both PaLM-E \cite{driess23palme} and RT-2 \cite{brohan23rt2} have achieved a tighter integration of perception and action planning, but still primarily focus on fundamental ``pick-and-place'' tasks or spatial rearrangements and inherently treat objects as passive targets without investigating into “tools”, which play important roles in complex tasks and plans. BEHAVIOR-1K \cite{li24behavior1k} challenges agents with realistic physics and demanding interaction with rigid bodies, deformable materials, and complex thermal states. Yet it did not explicitly assess the zero-shot cognitive capacity of multimodal foundation models to comprehend and plan with specialized equipment.

More recently, studies have explicitly begun to explore the intersection of LLMs and robotic tool use. For example, RoboTool \cite{xu24robotool} leverages a multi-agent LLM pipeline to generate executable code, enabling robots to utilize objects creatively to overcome implicit physical constraints. Furthermore, its evaluation is severely limited in scale, encompassing a mere six task scenarios, which falls drastically short of providing a comprehensive assessment of tool-use capabilities. Because these frameworks often bypass the raw visual perception challenge by relying on predefined states or simplified environments, they fundamentally fail to evaluate an agent's capability to visually recognize diverse, professional physical tools from complex real-world scenes.







\section{The Physical Tool Bench}

\begin{figure*}[t]
\centering
\includegraphics[width=\textwidth]{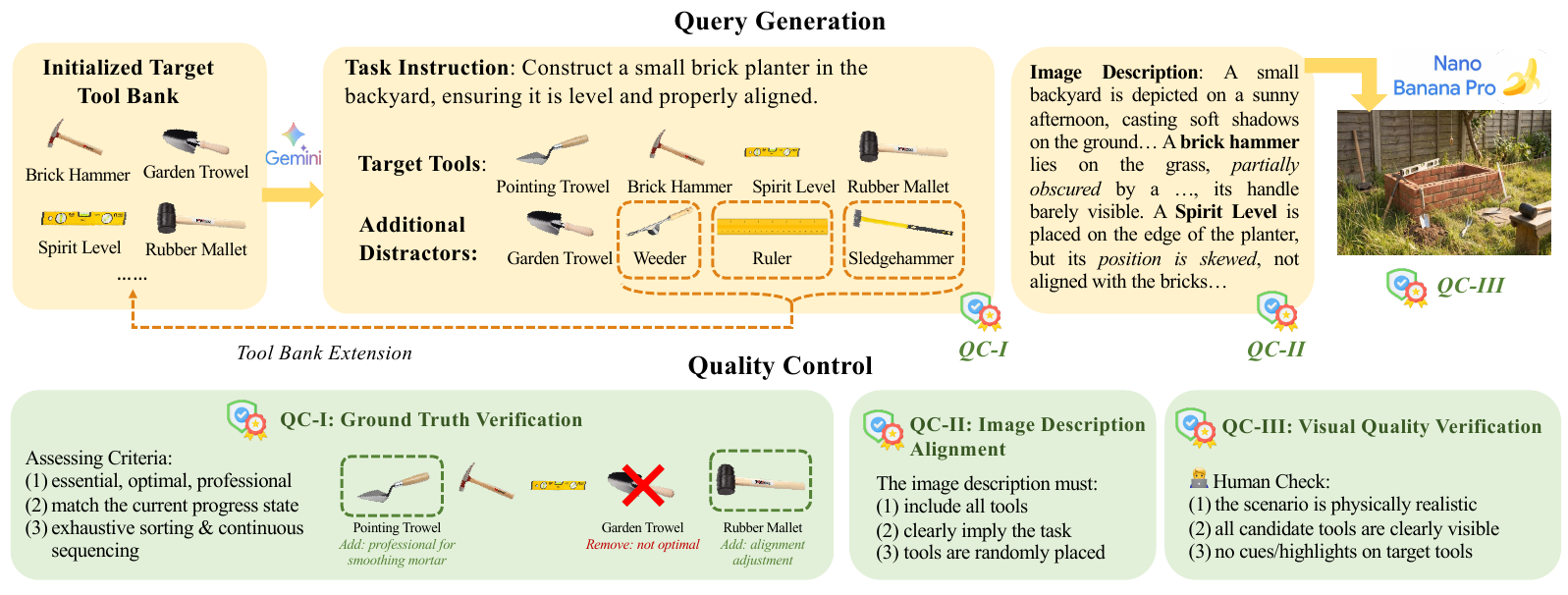}
\caption{Overview of the \textit{\benchname} construction pipeline. Gemini generates each query (task instruction, target tools, distractors) from the tool bank, with novel distractors recycled back via \emph{Tool Bank Extension}; Nano Banana Pro then renders the scene. Three quality-control stages follow: \textbf{QC-I} refines targets and assigns step labels; \textbf{QC-II} verifies tool-description alignment; \textbf{QC-III} applies human review for visual realism.}
\label{fig:fig_annotation_pipeline}
\end{figure*}

This section details the construction and characteristics of our proposed benchmark. We first outline the definitions of two primary tasks (\S~\ref{subsec:task_definition}). Next, we describe the annotation pipeline and quality assurance procedures for benchmark construction, which encompass target tool combination, instruction design, the injection of confounding tools, and the generation of visual scenarios (\S~\ref{subsec:dataset_pipeline}). Finally, we present an analysis of the dataset statistics (\S~\ref{subsec:statistics}).

\subsection{Problem Formulation}
\label{subsec:task_definition}

Each evaluation instance (a \emph{query}) is a tuple $(I, L)$, where $I$ is an image depicting a physical scenario with a set of available tools and $L$ is a natural language instruction (e.g., ``bond the cracked ceramic fragments''). Let $\mathcal{C} = \{c_1, \dots, c_N\}$ denote the complete set of tools visible in $I$, which includes both task-relevant targets and other items present in the scene. We evaluate MLLMs $f_{\theta}$ on two progressive tasks.

\heading{Task I: Physical Tool Recognition.}
Given the image $I$ and a recognition prompt $P_{rec}$, the model produces a predicted tool set $\hat{\mathcal{C}} = f_{\theta}(I, P_{rec})$, and the goal is to recover $\mathcal{C}$. This task isolates the model's ability to enumerate fine-grained physical tools from cluttered scenes, independent of any task instruction.

\heading{Task II: Tool Selection and Action Planning.}
Given $I$ and $L$, the model outputs an ordered sequence $\hat{Y} = f_{\theta}(I, L) = (y_1, \dots, y_K)$ with each $y_i \in \mathcal{C}$. The ground truth is $\mathcal{T}^* = \{(t_j, s_j)\}_{j=1}^{M}$, where $t_j \in \mathcal{C}$ and $s_j \in \mathbb{Z}_{\geq 1}$ is the execution-step index of $t_j$. Tools sharing the same $s$ are interchangeable, while tools with different $s$ values must follow their precedence ($s_j < s_k \Rightarrow t_j$ precedes $t_k$). This unifies ordered and order-free queries under one formulation. A prediction $\hat{Y}$ is correct iff (i) it forms a bijection with $\{t_j\}_{j=1}^{M}$ (no missing targets, no extras, no duplicates) and (ii) the step labels of its matched tools form a non-decreasing sequence.

\subsection{Dataset Construction Pipeline}
\label{subsec:dataset_pipeline}

Constructing \textit{\benchname} requires balancing two goals: covering the diversity of real-world physical tools while ensuring each query admits a clear, verifiable ground-truth solution. We achieve this through a three-stage pipeline (Figure~\ref{fig:fig_annotation_pipeline}): tool bank initialization (§\ref{subsubsec:tool_bank}), query generation (§\ref{subsubsec:query_generation}), and multi-stage quality assurance (§\ref{subsubsec:quality_control}). All the prompts utilized in this section are elaborated in Appendix~\ref{app:prompts}.

\subsubsection{Tool Bank Initialization and Extension}
\label{subsubsec:tool_bank}
We begin from a manually curated seed set of 310 commonly used physical tools and iteratively expand it during dataset construction along two complementary paths. First, we prompt the LLM to propose new tools across diverse application domains and functional categories, enforcing breadth across the bank. Second, novel distractor tools introduced by Gemini-3.1-Pro \cite{gemini25gemini} during query generation (§\ref{subsubsec:query_generation}) are recycled back into the bank (see the \emph{Tool Bank Extension} loop in Figure~\ref{fig:fig_annotation_pipeline}), systematically capturing functionally adjacent and visually similar confounders rather than only canonical task-completing tools. The expansion terminates at saturation, yielding 2,678 distinct tools.

\subsubsection{Query Generation}  
\label{subsubsec:query_generation}
\heading{Target tool combinations.} We prompt Gemini-3-Pro \cite{gemini25gemini} to formulate physical tool combinations restricted to the tool bank. Each query requires 1--3 target tools at generation time: 310 single-tool queries (one per tool, ensuring coverage), 500 two-tool queries, and 500 three-tool queries. When sampling these target tool combinations, we strictly control the selection frequency to prevent any specific tool from being overrepresented or underrepresented, thereby maintaining a balanced distribution.

\heading{Step labeling.} For multi-tool combinations, we also assign each tool an execution-step index $s_j$: tools that must be used before others receive earlier step indices, while functionally interchangeable tools share the same step. This step structure forms part of the ground truth $\mathcal{T}^*$ defined in §\ref{subsec:task_definition}.

\heading{Instruction and addition-tool injection.}
For each target combination, GPT-4o~\cite{openai24gpt4o} derives two distinct query scenarios. Instructions are phrased to describe the objective without naming required tools. Each scenario also receives 3--10 additional tools (i.e., distractors) selected for visual similarity, functional proximity, or domain relevance to the targets, reflecting the visual complexity of real-world environments.

\heading{Image description.} To prepare each query for visual rendering, we synthesize a detailed image description $d_i$ that specifies the scene composition. The description explicitly lists every candidate tool (both targets and additional tools) and instructs target tools to be randomly placed or partially obstructed to mimic real-world clutter.

\heading{Image rendering.} 
Using each description $d_i$, we render the corresponding scene image $I_i = \text{ImgGen}(d_i)$ with Nano Banana Pro\footnote{We additionally validate that our findings generalize beyond synthetically generated images on a real-world image subset in §\ref{sec:real_world}.}, supplemented with prompts enforcing adherence to physical laws.

\subsubsection{Multi-Stage Quality Assurance}
\label{subsubsec:quality_control}
To ensure the correctness of the ground truth and eliminate ambiguities, as illustrated in Figure~\ref{fig:fig_annotation_pipeline}, we implement three Quality Control (QC) checkpoints.

\heading{QC-I: Ground Truth Verification.} 
We refine each ground-truth target set with Gemini-3.1-Pro. Given the instruction, the scene description, and a shuffled list of all candidate tools, the model evaluates each tool against three criteria: (1) essential and professional for the task, (2) consistent with the scenario state, and (3) supporting a valid execution sequence. Based on this audit, tools may be reassigned between the target and distractor sets to eliminate cases where a distractor could substitute for a target.

\heading{QC-II: Image Description Alignment.} 
For each query, we run a programmatic check ensuring that every tool in the candidate set $\mathcal{C}$ appears as a literal mention in the image description $d_i$, preventing missing or hallucinated tools at rendering time.

\heading{QC-III: Visual Quality Verification.} 
Each rendered image undergoes a final human verification stage to filter out: (1) physically unrealistic scenarios, (2) images where candidate tools are not clearly visible, and (3) critically, images containing artificial cues such as unnatural highlighting or centralizing of target tools, which would allow models to bypass physical reasoning. After filtering, the final dataset contains 2,510 verified scenarios.

\subsection{Dataset Statistics}
\label{subsec:statistics}
The \textit{\benchname} encompasses 2,510 distinct evaluation scenarios over a diverse pool of 2,678 unique physical tools, comprising 1,168 target (positive) tools and 1,519 tools that only appear as confounders. All tools are classified into 57 segments based on the United Nations Standard Products and Services Code (UNSPSC), spanning manufacturing, electrical work, healthcare, agriculture, and beyond.
To evaluate resistance to visual distractors, each scenario presents a complex environment densely populated with candidate items, containing on average 8.62 tools (3.11 positive, 5.51 distractors). 86.9\% of scenarios require a strict sequential execution order, while the remainder evaluate order-free combinations. Query instructions are concise (avg.\ 103 characters), whereas the synthesized image descriptions used for rendering are highly detailed (avg.\ 1,736 characters), ensuring physical realism and exact alignment with the candidate tool constraints.
\section{Experimental Setup}

\begin{table*}[t]
    \centering
    \caption{Quantitative results on the proposed benchmark across various MLLMs. \textit{Order-Agnostic} reports Task I (visual recognition: identify every available tool in the image) with Precision, Recall, F1. \textit{Order-Aware} reports Task II (selection / planning) with Exact Match, Task-Completable Rate, Success Rate @ $k$. Subscripts on Task II cells are the Wilson 95\% confidence half-widths over the scenario sample. ``I'' and ``T'' denote the Instruct and Thinking model. Best results are bolded.}
    \label{tab:main_results}
    \newcommand{\ci}[1]{{\scriptsize #1}}
    \resizebox{\textwidth}{!}{
    \begin{tabular}{@{} l ccc
        r@{\,$\pm$\,}l r@{\,$\pm$\,}l r@{\,$\pm$\,}l r@{\,$\pm$\,}l r@{\,$\pm$\,}l @{}}
        \toprule
        & \multicolumn{3}{c}{\textbf{Order-Agnostic — Task I (\%)}}
        & \multicolumn{10}{c}{\textbf{Order-Aware — Task II (\%)}} \\
        \cmidrule(lr){2-4} \cmidrule(l){5-14}
        \textbf{Model} & \textbf{Precision} & \textbf{Recall} & \textbf{F1-score}
        & \multicolumn{2}{c}{\textbf{Overall EM}} & \multicolumn{2}{c}{\textbf{TCR}}
        & \multicolumn{2}{c}{\textbf{SR @ 1}} & \multicolumn{2}{c}{\textbf{SR @ 2}}
        & \multicolumn{2}{c}{\textbf{SR @ 3}} \\
        \midrule
        GPT-4o~\cite{openai24gpt4o} & \textbf{65.15} & 55.08 & 58.54
            &  5.62 & \ci{0.90} & 23.04 & \ci{1.65} & 38.53 & \ci{2.04} & 15.14 & \ci{1.50} &  3.99 & \ci{0.82} \\
        Qwen3-VL-Plus~\cite{bai25qwen3vl}           & 61.93 & \textbf{65.41} & \textbf{62.37}
            &  5.66 & \ci{0.91} & 20.81 & \ci{1.59} & 39.05 & \ci{2.05} & 16.52 & \ci{1.56} &  4.59 & \ci{0.88} \\
        GPT-5.2~\cite{openai26gpt52}    & 63.76 & 59.86 & 60.26
            & 10.66 & \ci{1.21} & 24.72 & \ci{1.69} & 47.59 & \ci{2.10} & 22.07 & \ci{1.74} &  6.80 & \ci{1.06} \\
        Gemini-3.1-Pro~\cite{gemini25gemini}    & 64.98 & 56.42 & 58.68
            & \textbf{20.96} & \ci{1.59} & \textbf{32.12} & \ci{1.83} & \textbf{55.83} & \ci{2.08} & \textbf{33.35} & \ci{1.98} & \textbf{13.90} & \ci{1.45} \\
        \cmidrule(l){2-14}
        Deepseek-VL2~\cite{wu24deepseekvl2}          & 51.31 & 43.74 & 44.48
            &  0.44 & \ci{0.27} & 12.48 & \ci{1.29} & 16.01 & \ci{1.54} &  4.50 & \ci{0.87} &  0.78 & \ci{0.38} \\
        MiniCPM      \cite{yu25minicpmv}           & 48.39 & 56.90 & 49.86
            &  1.00 & \ci{0.40} & 15.23 & \ci{1.41} & 26.24 & \ci{1.85} &  6.93 & \ci{1.07} &  1.79 & \ci{0.56} \\
        mPLUG-Owl3~\cite{ye24mplugowl3}            & 43.32 & 22.18 & 27.60
            &  1.12 & \ci{0.42} & 11.56 & \ci{1.25} & 16.97 & \ci{1.58} &  3.99 & \ci{0.82} &  0.73 & \ci{0.37} \\
        Qwen3-VL-32B-I \cite{bai25qwen3vl}  & 47.55 & 57.17 & 49.83
            &  1.24 & \ci{0.44} & 19.97 & \ci{1.56} & 30.46 & \ci{1.93} & 11.56 & \ci{1.34} &  3.07 & \ci{0.73} \\
        OpenFlamingo~\cite{awadalla23openflamingo}         & 19.48 & 19.79 & 18.37
            &  1.79 & \ci{0.52} &  3.59 & \ci{0.73} &  4.54 & \ci{0.88} &  0.69 & \ci{0.36} &  0.00 & \ci{0.09} \\
        InternVL3.5-38B~\cite{wang25internvl35}      & 50.87 & 42.41 & 44.70
            &  2.51 & \ci{0.62} & 13.71 & \ci{1.35} & 27.02 & \ci{1.86} &  8.67 & \ci{1.18} &  1.70 & \ci{0.55} \\
        OVis 2.6~\cite{lu24ovis}         & 64.83 & 49.25 & 53.18
            &  6.02 & \ci{0.93} & 15.46 & \ci{1.41} & 33.76 & \ci{1.98} & 12.57 & \ci{1.39} &  3.03 & \ci{0.72} \\
        Kimi-VL-A3B-T~\cite{bai26kimik25}   & 58.60 & 50.82 & 52.91
            &  6.78 & \ci{0.98} & 14.39 & \ci{1.37} & 31.56 & \ci{1.95} & 11.47 & \ci{1.34} &  2.61 & \ci{0.67} \\
        Qwen3-VL-32B-T~\cite{bai25qwen3vl}  & 64.16 & 47.87 & 53.15
            &  9.33 & \ci{1.14} & 18.17 & \ci{1.51} & 40.50 & \ci{2.06} & 16.79 & \ci{1.57} &  4.63 & \ci{0.89} \\
        \bottomrule
    \end{tabular}
    }
  \end{table*}

\label{sec:experimental_setup}
To establish a comprehensive baseline for physical tool use, we rigorously evaluate a suite of state-of-the-art Multimodal Large Language Models (MLLMs). This section details the selected models, the prompting strategies, and the specific metrics used to quantify performance across our two primary evaluation tasks.

\subsection{Implementation Details}
\label{subsec:evaluated_models}
We select a representative set of leading MLLMs, encompassing both proprietary (closed-source) and open-weight architectures. For proprietary models, we evaluate GPT-4o, GPT-5.2, Gemini 3.1 Pro, and Qwen3-VL-Plus~\cite{openai24gpt4o, openai26gpt52, gemini25gemini, bai25qwen3vl}. For open-weight models, we include Qwen3-VL, InternVL3.5, Kimi-VL, DeepSeek-VL, mPLUG-Owl3, OpenFlamingo, MiniCPM, and Ovis 2.6~\cite{bai25qwen3vl, wang25internvl35, bai26kimik25, wu24deepseekvl2, ye24mplugowl3, awadalla23openflamingo, yu25minicpmv, lu24ovis} to assess the capabilities of publicly available architectures. 

All evaluations are conducted in a \textit{zero-shot} setting to test the models' inherent physical reasoning and zero-shot generalization capabilities without relying on query-specific fine-tuning or few-shot demonstrations. To ensure standardized outputs, we utilize a standardized prompt template that instructs the models to first analyze the visual scene before outputting the required tool list or sequence. The prompt templates are exact in the Appendix.

\subsection{Evaluation Metrics}
\label{subsec:evaluation_metrics}

We define quantitative metrics for the two tasks formulated in \S~\ref{subsec:task_definition}. For Task~I, we evaluate the predicted tool set $\hat{\mathcal{C}}$ against the ground truth $\mathcal{C}$ using standard \textbf{Precision}, \textbf{Recall}, and \textbf{F1-score}. For Task~II, we evaluate both \emph{which} tools are selected and \emph{whether} they are arranged in the correct order.

\heading{Selection (Order-Agnostic).}
We apply Precision, Recall, and F1 to compare the predicted tool set against the ground-truth target set $\{t_j\}_{j=1}^{M}$, ignoring order. This isolates selection accuracy from sequential planning.

\heading{Exact Match (EM).}
EM is a strict criterion that requires a prediction to perfectly match the ground truth. A prediction scores 1 only if (i) its selected tools exactly match the target set $\{t_j\}$, with no missing or extra tools, and (ii) the tools appear in an order consistent with their step labels $s_j$, i.e., tools assigned to earlier steps precede those assigned to later steps. Tools sharing the same step may appear in any order. Any deviation yields a score of 0, and EM is reported as the average across all queries.

\heading{Task-Completable Rate (TCR).}
TCR relaxes EM by allowing additional tools beyond the ground truth. A prediction scores 1 if all target tools appear in a step-consistent order, even if extra unnecessary tools are included. TCR thus reflects whether an agent could still complete the task, while EM additionally requires a \emph{minimal} plan.

\heading{Success Rate @ $k$ (SR@$k$).} SR@$k$ ($k \in \{1, 2, 3\}$) measures EM restricted to the first $k$ tools in the predicted sequence. SR@$k$ captures how early in the sequence a model begins to fail and complements the all-or-nothing nature of EM.
\section{Results and Analysis}
\label{sec:results}

We empirically evaluate the suite of MLLMs introduced in §\ref{sec:experimental_setup} on our benchmark. Our analysis proceeds from overall performance (§\ref{sec:main_results}) to fine-grained breakdowns (§\ref{sec:fine_grained}), targeted probing studies (§\ref{sec:probing}), validation on real-world images (§\ref{sec:real_world}), and a fine-grained error analysis (§\ref{sec:errors}).

\subsection{Main Results}
\label{sec:main_results}
Table~\ref{tab:main_results} reports overall performance across all evaluated MLLMs on both Task~I (Physical Tool Recognition over all available tools in the scene) and Task~II (Tool Selection and Action Planning conditioned on the task instruction). Three findings stand out.

\paragraph{Recognition is non-trivial, even for SOTA models.} When asked to enumerate every tool visible in a real scene (Task~I), no model exceeds 63\% F1: the best score is Qwen3-VL-Plus at 62.37\%, and the majority fall below 50\%. Smaller open-weight models such as mPLUG-Owl3 (27.60\%) and OpenFlamingo (18.37\%) miss more than 70\% of the tools present. Adding the task instruction (Task~II) does not consistently help: only 4 of 13 models improve over their Task~I F1, with the rest performing comparably or worse (full comparison in Appendix~\ref{appx:f1_comparison_tasks}). This is because Task~II requires not only perceiving the tools, but reasoning about their \emph{functional relevance} to the instruction. Many MLLMs recognize tools in Task~I yet fail to map them onto task semantics in Task~II, pointing to a more cognitive bottleneck that we examine in §\ref{sec:probing}.

\paragraph{A large gap separates recognition from planning.} Despite the perceptual advantage afforded by task instructions, the highest overall Exact Match (EM) on Task~II is only 20.96\% (Gemini-3.1-Pro). The order-aware metrics deteriorate even more sharply: the best Success Rate at $k{=}3$ is 13.90\% (Gemini-3.1-Pro), and no model exceeds 56\% even at $k{=}1$. This decoupling suggests that current MLLMs may \emph{see} the right tools without being able to reason about which subset to use, in what order, and why.

\paragraph{Closed-source models lead, but the gap is narrowing.} Proprietary models (GPT-4o, GPT-5.2, Gemini-3.1-Pro) consistently outperform their open-source counterparts on Task~II EM, with Gemini-3.1-Pro leading on every order-aware metric. Nevertheless, the strongest open-source reasoning models, Qwen3-VL-32B-Thinking (9.33\% EM) and Kimi-VL-A3B-Thinking (6.78\% EM), match or exceed GPT-4o (5.62\% EM) on several order-aware metrics, narrowing the gap on planning-style tasks.

\subsection{Fine-grained Analysis}
\label{sec:fine_grained}

\subsubsection{Effect of Query Complexity}
\label{sec:complexity}

\begin{figure}[h]
    \centering
    \includegraphics[width=\linewidth]{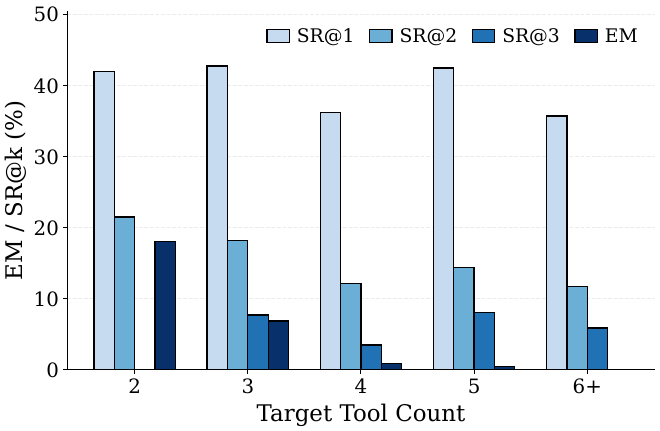}
    \caption{Exact-match performance of Qwen3-VL-32B-Thinking on Task~II across the number of target tools $k$. SR@$j$ requires the first $j$ predicted tools to match the ground truth prefix; EM additionally forbids extra tools beyond the ground truth. SR@3 is undefined when $k{=}2$. While SR@1 stays at 54--57\% across all complexities, EM collapses from 34.5\% at $k{=}2$ to 0.5\% at $k{=}6{+}$, exposing a sharp degradation in multi-step planning.
    }
    \label{fig:task_ii_results}
\end{figure}

Figure~\ref{fig:task_ii_results} presents Gemini-3.1-Pro's Task~II performance by the number of target tools. SR@1 remains nearly constant across complexities (54--57\%), indicating that selecting the first appropriate tool is largely insensitive to query length. In sharp contrast, EM collapses from 34.5\% at $k{=}2$ to 0.5\% at $k{=}6{+}$, with SR@3 falling below 20\% once the query requires four or more tools. The widening gap between SR@1 and EM thus directly quantifies the model's failure to maintain a globally consistent execution plan: even when individual tools are correctly identified at the start, the probability of completing the full sequence decays super-linearly with complexity. This pattern indicates that the dominant source of difficulty is multi-step physical planning rather than single-step tool recognition.

\subsubsection{Performance Across UNSPSC Domains}
\label{sec:domain}
We further disaggregate Task~II EM across the seven broad UNSPSC domains (see Appendix~\ref{appx:per_category} for the full breakdown). Models perform substantially better on \emph{Healthcare} and \emph{Office} scenarios, where procedures are well-defined and tool sets are small, but degrade markedly on \emph{Manufacturing} and \emph{Electrical Work}, where ordering constraints are strict and confounding tools share both visual and functional similarities. This pattern points to a systematic deficit in domain-specific physical commonsense rather than a uniform recognition limitation.

\subsection{Probing Studies}
\label{sec:probing}

\subsubsection{Perception Ceiling}
\label{sec:perception_ceiling}
To localize the MLLM perception bottleneck, we evaluate state-of-the-art open-vocabulary object detectors on the same scenes. Given the candidate tool list as a text prompt, Grounding DINO achieves a recall of 70.53\% --- exceeding the best MLLM (Gemini-3.1-Pro at 57.09\% on Task~I) by 13.44 percentage points. This indicates that the visual evidence required for tool recognition is present in the images, and MLLM failures are not driven by raw perception but by the inability to enumerate visible tools or to ground them in the task instruction.

\subsubsection{Human Reference}
\label{sec:human_baseline}

To contextualize model performance, one annotator from our research team completed a stratified sample of 100 queries, rating their domain familiarity from 1 to 5 per query. On items rated highly familiar (confidence~5), the annotator achieves \textbf{75\% EM}, \textbf{75\% TCR}, and \textbf{95\% F1}, indicating that the benchmark admits well-defined answers aligned with informed human judgment. Across all familiarity levels, the annotator reaches \textbf{38\% EM}, \textbf{49\% TCR}, and \textbf{80.6\% F1}, still substantially exceeding the best MLLM (Gemini-3.1-Pro at 21.0\% EM). The model deficits thus reflect capability limitations rather than task ambiguity. We leave a multi-annotator study to future work.

\subsection{Real-World Image Validation}
\label{sec:real_world}
A natural concern is whether our findings generalize beyond synthetically generated images. To address this, we construct a real-world image subset of 201 queries collected from web sources, manually matching the images to task instructions from the benchmark while preserving the original target labels.

Base on the evaluated results, precision drops by 8.95 percentage points on the real-world subset, whereas EM remains nearly unchanged (19.9\% on generated images vs.\ 19.4\% on real-world images). The degradation in the order-agnostic metrics is consistent with the lower image quality of in-the-wild photographs, which exhibit varied resolution, lighting conditions, and motion blur. These results indicate that synthetic image generation, provides a charitable testbed: the capability gap exposed by our benchmark is not an artifact of the synthetic distribution and would likely be more pronounced under real-world deployment.

\subsection{Error Analysis}
\label{sec:errors}

\begin{figure}[h]
    \centering
    \includegraphics[width=\linewidth]{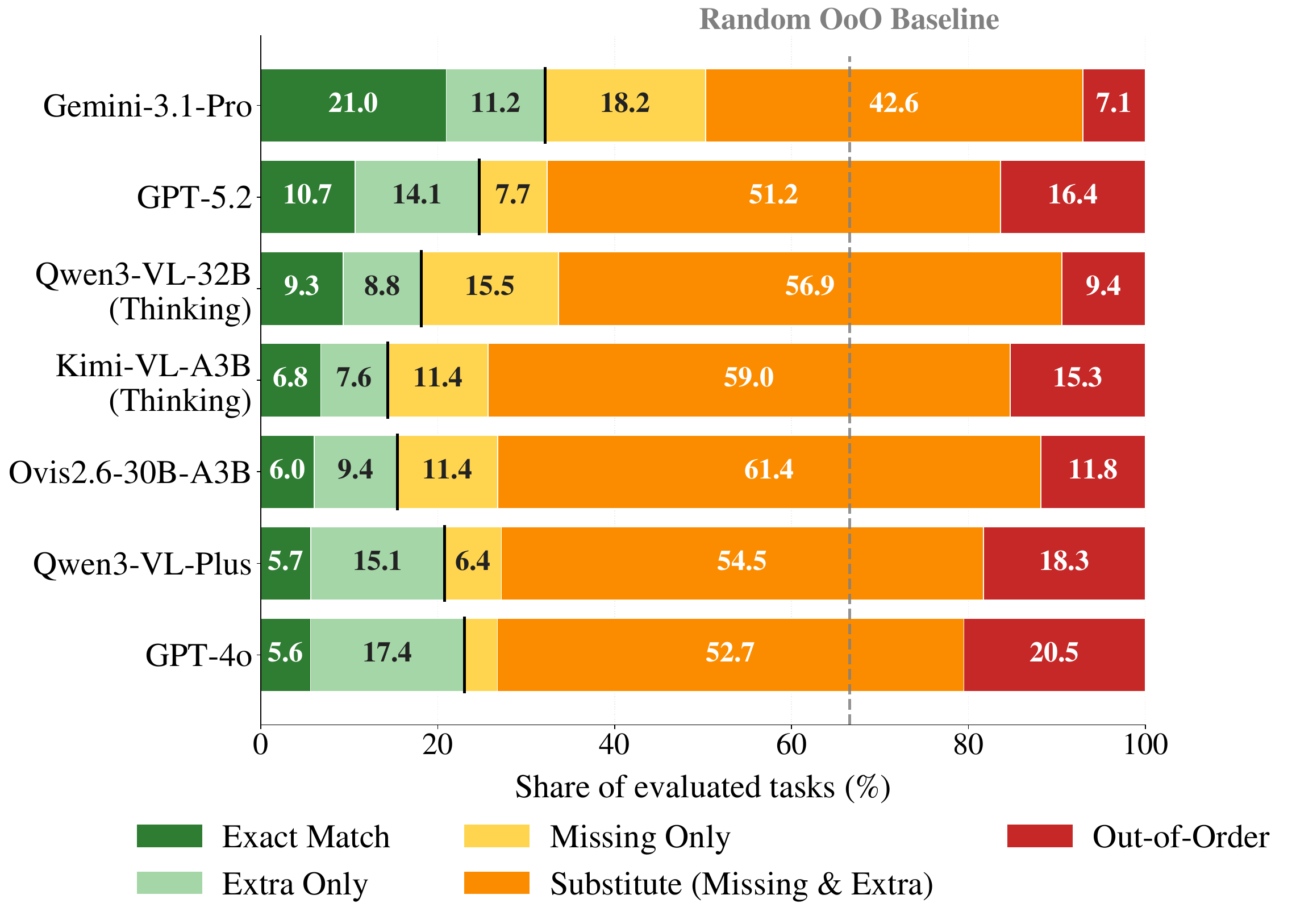}
    \caption{Task~II failure decomposition across seven representative MLLMs, sorted by Exact Match (EM). Each prediction is assigned to one of five mutually exclusive outcomes. The first two (EM, Extra Only) are task-completable; the remaining three are task-blocking. The dashed line marks the expected Out-of-Order rate under random tool selection and ordering (33.5\%).}
    \label{fig:error_profile}
\end{figure}

Figure~\ref{fig:error_profile} decomposes each Task~II prediction into five mutually exclusive outcomes, of which the first two (Exact Match, Extra Only) are \textbf{task-completable} and the remaining three are \textbf{task-blocking}. To probe the underlying causes, we additionally annotate 100 failure cases from Gemini-3.1-Pro. Three observations stand out, with qualitative examples for each error category provided in Appendix~\ref{app:qualitative}.

\heading{Substitution dominates, with functional confusion as the primary driver.} As shown in Figure~\ref{fig:error_profile}, Substitute---where at least one target tool is replaced by a distractor---is the largest failure mode for every model. Our manual annotation reveals that the missing-target component of these failures is rarely caused by perception: only 22\% of missed tools are visually occluded or too small to recognize, while \textbf{41.3\% are \emph{functional omissions}}---tools correctly identified in Task~I but excluded from the Task~II plan because the model fails to recognize their functional relevance to the instruction. A further 36.7\% are tools clearly visible in the scene but not recognized in either task, reflecting a fine-grained recognition gap rather than visual difficulty. On the spurious-selection side, 60\% of incorrectly selected tools are distractors actually present in the scene, and 40\% are hallucinated tools not visible at all. Together, these results indicate that the bottleneck is task-conditioned functional reasoning rather than raw perception.

\heading{Ordering competence exists but is fragile.} Out-of-Order rates (Figure~\ref{fig:error_profile}) sit well below the 33.5\% random baseline, indicating non-trivial sequencing ability. However, root-cause analysis shows that 50\% of OoO failures stem from misinterpreting the task instruction rather than generic ordering weakness, suggesting that improving instruction grounding may directly reduce ordering errors.

\heading{Failure profiles diverge across model families.} Thinking models (Qwen3-VL-32B-Thinking, Kimi-VL-A3B-Thinking) trade lower ordering errors for higher Substitute rates, while GPT-4o and Qwen3-VL-Plus show the opposite pattern---high OoO with comparatively lower Substitute. Explicit reasoning thus improves sequential planning yet leaves the functional-disambiguation gap untouched. These contrasts are invisible at the aggregate EM level but become apparent in the per-category decomposition shown in Figure~\ref{fig:error_profile}.
\section{Conclusion}
We introduce \textit{\benchname}, the first benchmark dedicated to evaluating physical tool use in MLLMs. Across 13 leading models, we find a substantial gap between digital and physical tool use: even the strongest MLLMs complete only a small fraction of queries end-to-end, and most failures arise from substituting target tools with functionally similar alternatives that are visible in the scene.

This bottleneck is not raw perception. Specialized detectors and humans both substantially outperform current MLLMs, and recognition recall persists on real-world images. The deficit lies in the functional commonsense required to map perceived tools onto task semantics. Closing this gap is unlikely to come from scaling visual encoders alone; we believe progress will require explicit grounding in multi-step physical reasoning, particularly for the long tail of specialized domains where embodied AI is most likely to be deployed.

\section*{Limitations}

\heading{Coverage of tool categories.} While \textit{\benchname} spans 57 UNSPSC segments and 2{,}678 distinct tools, certain specialized domains are underrepresented due to the difficulty of obtaining realistic visual references. Expanding to these long-tail domains is a natural direction for future iterations.

\paragraph{Static visual contexts.}
\textit{\benchname} evaluates tool use from a single static scene image, without modeling dynamic state changes (e.g., the workpiece evolving as it is processed) or interactive feedback (e.g., the model querying additional viewpoints). Extending to multi-turn, interactive tool-use evaluation is a promising direction for future work.

\section*{Ethical Considerations}

\heading{Data sources and licensing.} Synthetic images were generated by Nano Banana Pro in compliance with its terms of service. The real-world image subset (§\ref{sec:real_world}) was collected from publicly available web sources under fair use for academic research; 

\heading{Model access.} All evaluated proprietary models were accessed through their official APIs in accordance with each provider's terms of use.

\heading{Human reference.} The human reference (§\ref{sec:human_baseline}) and QC-III visual verification were conducted by a research team member who consented to the task and were informed of the research purpose. The task involves only assessing visual scenes of physical tools and contains no personally identifiable information or sensitive content.


\bibliographystyle{unsrtnat} 
\bibliography{ref}


\appendix
\begin{figure*}[t]
    \centering
    \includegraphics[width=\textwidth]{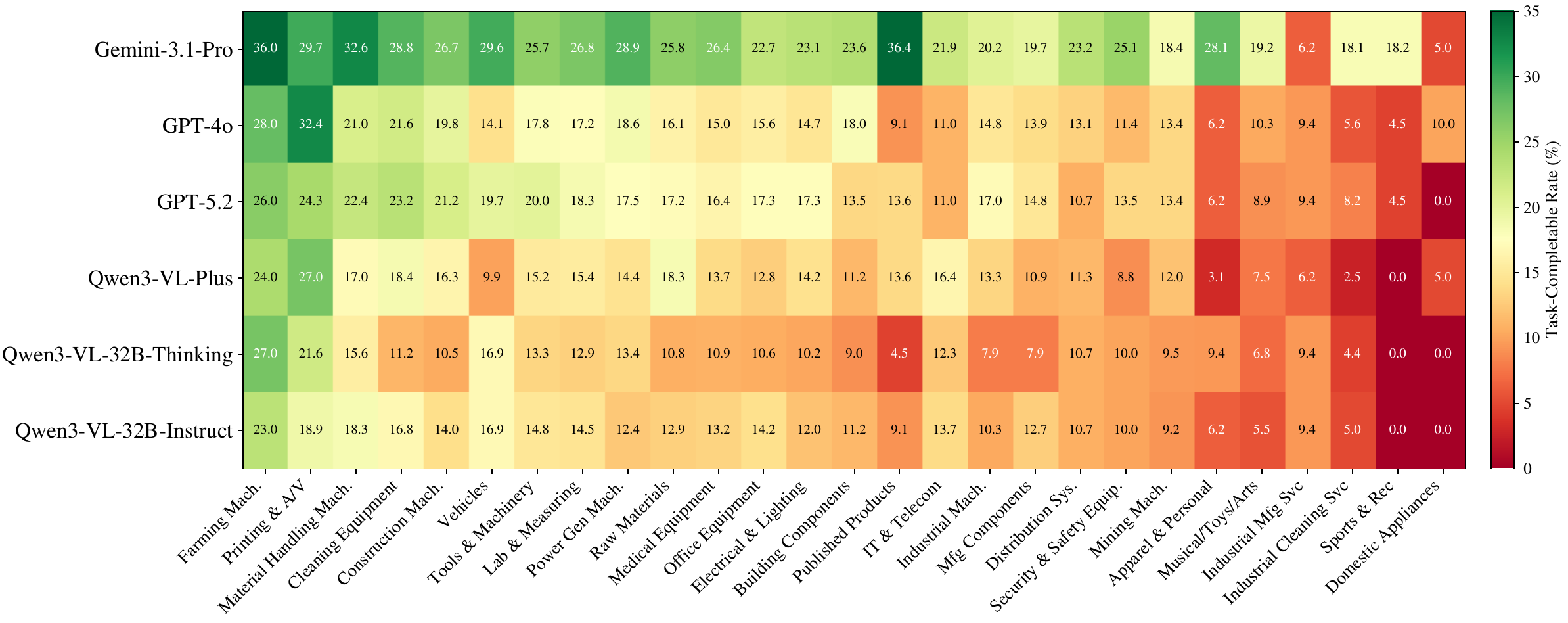}
    \caption{Per-segment Task-Completable Rate (TCR, \%) across 28 UNSPSC functional segments, for six representative MLLMs. Segments are sorted left-to-right by mean TCR across models (descending). All models exhibit a consistent decline from left to right, with the cross-model ranking of segments highly correlated, indicating that category-level difficulty is intrinsic to the task rather than model-specific.}
    \label{fig:per_category_tcr}
\end{figure*}

\begin{table*}[t]
    \centering
    \caption{Comparison of Task I (visual recognition: identify every tool visible in the image) and Task II (tool selection / planning) on the same 2{,}510 scenarios. Task I GT =
  \texttt{shuffled\_available\_tools}; Task II GT = task-relevant target tools. $\Delta$F1 = Task II F1 $-$ Task I F1 — positive values indicate the model is stronger at closed-set
   selection than open-set recognition. Best results per column are bolded.}
    \label{tab:f1_comparison_tasks}
    \resizebox{0.83475\width}{!}{
    \begin{tabular}{@{} l ccc ccc c @{}}
      \toprule
      & \multicolumn{3}{c}{\textbf{Task I — Recognition (\%)}} & \multicolumn{3}{c}{\textbf{Task II — Selection (\%)}} & \\
      \cmidrule(lr){2-4} \cmidrule(lr){5-7}
      \textbf{Model} & \textbf{Precision} & \textbf{Recall} & \textbf{F1} & \textbf{Precision} & \textbf{Recall} & \textbf{F1} & \textbf{$\Delta$F1} \\
      \midrule
      GPT-4o~\cite{openai24gpt4o}  & \textbf{65.15} & 55.08 & 58.54 & 53.56 & 63.90 & 55.86 & $-2.68$ \\
      Qwen3-VL-Plus                & 61.93 & \textbf{65.41} & \textbf{62.37} & 56.78 & 61.04 & 56.26 & $-6.11$ \\
      GPT-5.2                      & 63.76 & 59.86 & 60.26 & 59.44 & 64.82 & 59.50 & $-0.76$ \\
      Gemini-3.1-Pro               & 64.98 & 56.42 & 58.68 & \textbf{71.61} & \textbf{67.87} & \textbf{67.32} & $\mathbf{+8.64}$ \\
      \cmidrule(l){2-8}
      Deepseek-VL2                 & 51.31 & 43.74 & 44.48 & 34.86 & 51.03 & 39.34 & $-5.14$ \\
      MiniCPM                      & 48.39 & 56.90 & 49.86 & 37.96 & 54.45 & 42.25 & $-7.61$ \\
      mPLUG-Owl3                   & 43.32 & 22.18 & 27.60 & 29.28 & 50.09 & 33.35 & $+5.75$ \\
      Qwen3-VL-32B-Instruct        & 47.55 & 57.17 & 49.83 & 40.81 & 63.31 & 47.32 & $-2.51$ \\
      OpenFlamingo                 & 19.48 & 19.79 & 18.37 & 15.37 & 11.42 & 12.05 & $-6.32$ \\
      InternVL3.5-38B              & 50.87 & 42.41 & 44.70 & 46.89 & 49.64 & 45.99 & $+1.29$ \\
      OVis 2.6                     & 64.83 & 49.25 & 53.18 & 54.66 & 50.26 & 49.64 & $-3.54$ \\
      Kimi-VL-A3B-Thinking         & 58.60 & 50.82 & 52.91 & 54.17 & 49.80 & 49.11 & $-3.80$ \\
      Qwen3-VL-32B-Thinking        & 64.16 & 47.87 & 53.15 & 60.86 & 53.53 & 54.39 & $+1.24$ \\
      \bottomrule
    \end{tabular}
    }
  \end{table*}

\section{Per-Category Performance Analysis}
\label{appx:per_category}

To understand whether MLLMs exhibit uniform competence in physical tool use or whether their performance varies by tool category, we disaggregate the Task-Completable Rate (TCR) across the 28 UNSPSC functional segments covered by \textit{\benchname}. Figure~\ref{fig:per_category_tcr} reports the TCR of six representative MLLMs (Gemini-3.1-Pro, GPT-4o, GPT-5.2, Qwen3-VL-Plus, Qwen3-VL-32B-Thinking, and Qwen3-VL-32B-Instruct), with segments sorted by mean score across models.

\paragraph{Overall trend.} Performance varies dramatically across categories, spanning from above 30\% TCR on the easiest segments to near-zero on the hardest. Gemini-3.1-Pro maintains the highest TCR on the leftmost segments (e.g., 30.8\% on Farming Machinery, 27.5\% on Cleaning Equipment) but collapses on the rightmost categories, falling to 4.8--9.1\% on segments such as Industrial Cleaning Services, Sports \& Recreation, and Electronic Components. This pattern is consistent across all six evaluated models, with cross-model correlation of segment rankings exceeding 0.85: the same segments are easy or hard for every model.

\paragraph{What makes a category easy or hard?} The leftmost (easiest) segments\,---\,Farming Machinery, Cleaning Equipment, Construction Machinery, Vehicles, Power Generation\,---\,share two properties: (i) tools are \emph{visually distinctive} (e.g., a tractor or a leaf blower is unlikely to be confused with another tool), and (ii) the mapping from instruction to tool is largely \emph{one-to-one} (e.g., ``mow the lawn'' unambiguously implies a lawn mower). Under these conditions, even moderate functional reasoning suffices to recover the correct tool set.

In contrast, the rightmost (hardest) segments\,---\,Industrial Cleaning Services, Sports \& Recreation, Apparel \& Personal, Electronic Components\,---\,exhibit \emph{fine-grained functional overlap} among candidates. For instance, distinguishing between an arbor press, a hydraulic press, and a punch press in Industrial Mfg Services requires specialized domain knowledge that current MLLMs do not reliably possess. Similarly, Electronic Components scenarios frequently demand multi-step procedures involving visually similar instruments (e.g., a multimeter, an oscilloscope probe, and a logic analyzer), which our analysis in §\ref{sec:errors} identifies as a primary trigger of functional substitution errors.

\section{Task I vs.\ Task II: Recognition Under Instruction Conditioning}
\label{appx:f1_comparison_tasks}

To better understand whether task instructions help MLLMs identify relevant tools, we compare model performance on Task~I (visual recognition: identify every tool visible in the image) and Task~II (selection: identify only the tools required by the instruction) on the same 2{,}510 scenarios. Both tasks are evaluated with set-level Precision, Recall, and F1, computed against their respective ground-truth sets: Task~I against the full set of available tools in the scene, and Task~II against the task-relevant target tools. Table~\ref{tab:f1_comparison_tasks} reports the per-model breakdown along with $\Delta\text{F1} = \text{Task~II F1} - \text{Task~I F1}$, where positive values indicate that the model benefits from instruction conditioning.

\paragraph{Aggregate trend.} Across 13 evaluated models, only 5 exhibit a positive $\Delta\text{F1}$, while the remaining 8 perform comparably or worse on Task~II. The largest gains are observed on OVis~2.6 ($+10.80$\,pp) and Gemini-3.1-Pro ($+8.64$\,pp), suggesting that these models are able to leverage the instruction as an effective attentional prior. In contrast, most other models, including GPT-4o ($-2.68$\,pp), GPT-5.2 ($-0.76$\,pp), Qwen3-VL-Plus ($-6.11$\,pp), and several open-source models, show no improvement or a slight degradation.

\paragraph{Why does instruction conditioning not uniformly help?} The result is initially counterintuitive: one might expect the instruction to narrow the model's attention to a smaller, task-relevant subset of tools and thereby simplify the problem. However, Task~II imposes an additional reasoning demand on top of perception. The model must not only \emph{see} the tools, but also judge their \emph{functional relevance} to the instruction (e.g., recognizing that epoxy resin, rather than duct tape, is the appropriate adhesive for repairing ceramic). For models that lack robust physical commonsense, this additional reasoning step introduces errors that outweigh the benefit of a narrower target set: they may drop correct targets whose relevance is not obvious, or substitute them with functionally adjacent alternatives. Stronger models such as Gemini-3.1-Pro and OVis~2.6 appear better able to exploit the instruction without incurring these costs, whereas others are essentially neutralized or pulled down by the added cognitive burden.

\paragraph{Implications.} This pattern reinforces our central diagnosis: the bottleneck in physical tool use is not raw visual perception, but the higher-level reasoning required to ground perceived tools in task semantics. We provide complementary evidence for this view through a perception-ceiling experiment with open-vocabulary detectors (§\ref{sec:probing}) and an error decomposition that identifies functional substitution as the dominant failure mode (§\ref{sec:errors}).

\section{Prompt Templates}
\label{app:prompts}

This appendix lists the prompt templates used throughout the dataset construction pipeline (Sections~\ref{app:prompt_combination}--\ref{app:prompt_qc1}) and the evaluation procedure (Sections~\ref{app:prompt_task1}--\ref{app:prompt_judge}). For brevity, system messages and minor formatting tokens are omitted; full versions are released with the dataset.

\subsection{Target Tool Combination Generation}
\label{app:prompt_combination}
We further expand the tool bank by prompting Gemini-3.1-Pro to include common combination of 2 or 3 tools that usually worked together to complete a task. Notably, the task instructions generated here only serve as a guide to ensure the tool combination is feasible and commonly acknowledged in real-world, avoiding getting random combinations. These task instructions are dropped after we obtained the tool combinations.

\begin{tcolorbox}[colback=gray!5,colframe=gray!50,title=Prompt: Tool Combination Generation,breakable,enhanced]
\small\ttfamily
[You are an expert in tool selection and tool usage across diverse real-world domains. I have attached a set of tools. Your goal is to propose 100 distinct combinations of exactly 2 tools from this set. For each combination, design a specific, realistic target task that requires the usage of all tools to be successfully completed. For each combination, output a single JSON object containing exactly the following two fields: 
(1) task\_instruct: A clear task instruction written in English. The task must require the use of all the 2 target tools to be completed. Do NOT mention or imply any specific tools, including any target tools listed in tools\_target in (2). 
(2) tools\_target: 2 required tools needed to complete the task. The tools must be exactly from the attached tool list. If the tools are used in a specific order, list them in the correct operational sequence.]
\end{tcolorbox}

\subsection{Task Instruction Generation}
\label{app:prompt_instruction}
The task instruction is generated by prompting GPT-4o with the pre-determined initial target tools (in form of single tool or tool combination).
\begin{tcolorbox}[colback=gray!5,colframe=gray!50,title=Prompt: Task Instruction Generation]
\small\ttfamily
["task\_instruct": A clear task instruction in English. The task must require ALL the target tools [{tools\_list}] to be completed. Do NOT mention or imply any specific tool or contain part of the tool name word.]
\end{tcolorbox}

\subsection{Distractor Selection}
\label{app:prompt_distractor}
The distractors are selected by prompting GPT-4o with the target tool, along with the generation of task instruction and image descriptions. For each initial target tool(s), two distinct task scenarios will be constructed, and the two different numbers of distractors to include in each task scenario are randomly chosen between 3 and 10.
\begin{tcolorbox}[colback=gray!5,colframe=gray!50,title=Prompt: Distractor Selection]
\small\ttfamily
["tools\_negative": A list of tools that are NOT required for this task. 
   - Scenario 1 must have exactly {neg\_counts[0]} items.
   - Scenario 2 must have exactly {neg\_counts[1]} items.
These tools should be confusing or misleading - they might:
   - Look similar to the target tools
   - Have similar functions to the target tools
   - Be used on similar objects but be wrong choices
   - Be commonly associated with the same work domain
   Make these negative tools realistic distractors.]

\end{tcolorbox}

\subsection{Image Description Generation}
\label{app:prompt_imgdesc}
The image description is generated simultaneously with the task instruction and distractors by prompting GPT-4o. Before a piece of task scenario (including target tool(s), task instruction, distractors and image description) is saved, we would check and ensure that all the tools are clearly mentioned and addressed in the corresponding image description.
\begin{tcolorbox}[colback=gray!5,colframe=gray!50,title=Prompt: Image Description for nano-banana-pro]
\small\ttfamily
["img\_desc": A detailed English description of a single image depicting the scenario. The image must:
   - Clearly imply the task to be completed
   - Show ALL tools from both tools\_target and tools\_negative
   - Make the correct target tools look randomly placed and partially hidden; they should NOT be highlighted, should not be placed conspicuously, and should not appear ready to complete the task.
   - When describing technical or professional workspaces, ensure that tools adhere to their mechanical function. 
   - Include specific details about environment, lighting, angles, tool placement, and scene context
   - Be detailed enough to generate a realistic, plausible image. ]
\end{tcolorbox}

\subsection{QC-I: Target Tool Verification}
\label{app:prompt_qc1}
In QC-I, we refine the target tools and determine the chronological step orders in a more rigorous way by prompt Gemini-3.1-Pro with 'Task Instruction' , 'Current Scene' (the first paragraph of the image description), and 'Available Tools' (the combined set of initial target tool and distractors).
\begin{tcolorbox}[colback=gray!5,colframe=gray!50,title=Prompt: QC-I Target/Distractor Audit,breakable,enhanced]
\small\ttfamily
[You are an expert AI agent orchestrator evaluating tool selection capabilities across diverse professional domains. 
    I will provide a 'Task Instruction', a 'Current Scene' description, and 'Available Tools'.
    Your objective is to identify the ABSOLUTE MINIMAL REQUIRED SET of professional tools and sequence them based on scene progress.

    THE THREE LAWS OF TOOL ORCHESTRATION:
    1. THE UNIFIED VIABILITY TEST: A tool is strictly REQUIRED only if its removal causes the task to physically fail, violate safety, or violate professional industry standards.
       - Implicit Constraints: You must consider implicit constraints. (e.g., studying animals "without disturbing habitat" standardly requires an unattended tool like a 'Wildlife Camera Trap' to avoid human presence, making it professional necessity).
       - Technical Standards: You must prioritize professional-grade methods over amateur workarounds (e.g., prefer 'Heat Gun' over 'Electrical Tape' for professional automotive wiring).
       - Nice-to-Haves: Reject any tool that merely provides convenience but isn't required for success (e.g., GPS, Tripods).
    2. SHARP REDUNDANCY ELIMINATION: If multiple tools overlap in fulfilling the requirement of Law 1 (e.g., Telescope vs. Field Binoculars for mobile observation), you MUST select ONLY the single most contextually appropriate tool and move all alternatives and their specific accessories to 'negative\_tools'.
    3. TASK LIFECYCLE TRACKING: Evaluate `<img\_desc>` to determine what has already been completed.
       - REJECT tools meant ONLY for phases already finished in the image.
       - RETAIN and DELAY tools needed for remaining phases, final reassembly, or closing up to the LAST steps of the sequence.

    Rules for Output (Strictly Follow JSON Schema):
    1. 'tool\_analysis': Step-by-step evaluation of EVERY available tool.
       - 'viability\_and\_standard\_justification': Explain why this tool is a professional and physical necessity based on Law 1 and 2. Write 'Failed' if it is non-essential, amateurish, or redundant.
       - 'status': "Target" OR "Negative" (state exact reason: Non-Essential / Substandard / Redundant / Already Completed).
       - 'sequence\_logic': Timing rationale based on scene progress, or 'None'.
    2. 'target\_tools': List of selected tools.
    3. 'target\_steps': Integers representing the execution order (starting at 1, continuous, same number for parallel tools).
    4. 'negative\_tools': List of rejected tools.
]
\end{tcolorbox}

\subsection{Evaluation Prompt --- Task~I (Tool Recognition)}
\label{app:prompt_task1}
We test MLLM's ability in recognizing all available tools in the scene by the following prompt.
\begin{tcolorbox}[colback=gray!5,colframe=gray!50,title=Prompt: Task I --- Tool Recognition]
\small\ttfamily
[List all tools in this image. Please provide only the names of the tools, separated by commas. Do not include any explanations or extra text.]
\end{tcolorbox}

\subsection{Evaluation Prompt --- Task~II (Tool Selection and Planning)}
\label{app:prompt_task2}
We further evaluate MLLM's ability to address the task in the provided scene by the following prompt.
\begin{tcolorbox}[colback=gray!5,colframe=gray!50,title=Prompt: Task II --- Tool Selection and Action Planning]
\small\ttfamily
[Given the following TASK, which tool(s) in the image are most appropriate to complete the task? Please list the name(s) of the selected tools in the order they should be used and separate them by commas. No explanation needed. TASK: {task\_instruct}. SELECTED TOOL(S) (in order of use):]
\end{tcolorbox}

\subsection{LLM-as-Judge Prompt}
\label{app:prompt_judge}

To match model predictions against the ground truth, we employ a hybrid pipeline combining case-insensitive string matching with the following LLM judge.

\begin{tcolorbox}[colback=gray!5,colframe=gray!50,title=Prompt: LLM-as-Judge]
\small\ttfamily
[You are an expert evaluator. I have a list of 'Identified Tools' predicted by a model. 
    Your task is to map each 'Identified Tool' to the correct 'Target Tool' name (if applicable) for the provided task, while ensuring it does not refer to any 'Negative Tools' (distractors).

    Rules:
    1. Only match if the Identified Tool is clearly the same tool as a Target Tool.
    2. If the Identified Tool is ambiguous and could potentially refer to a Negative Tool, DO NOT match it.
    3. Use the exact string from the Target Tools list for the value in your mapping.
    4. Return ONLY a valid JSON object where keys are the Identified Tool strings and values are the corresponding Target Tool strings.
    5. DO NOT map multiple Identified Tools to the same Target Tool – each target can appear at most once.]
\end{tcolorbox}

\section{Qualitative Examples}
\label{app:qualitative}

\subsection{Successful Cases}
\label{app:success_cases}
\begin{figure}[t]
    \centering
    \includegraphics[width=\linewidth]{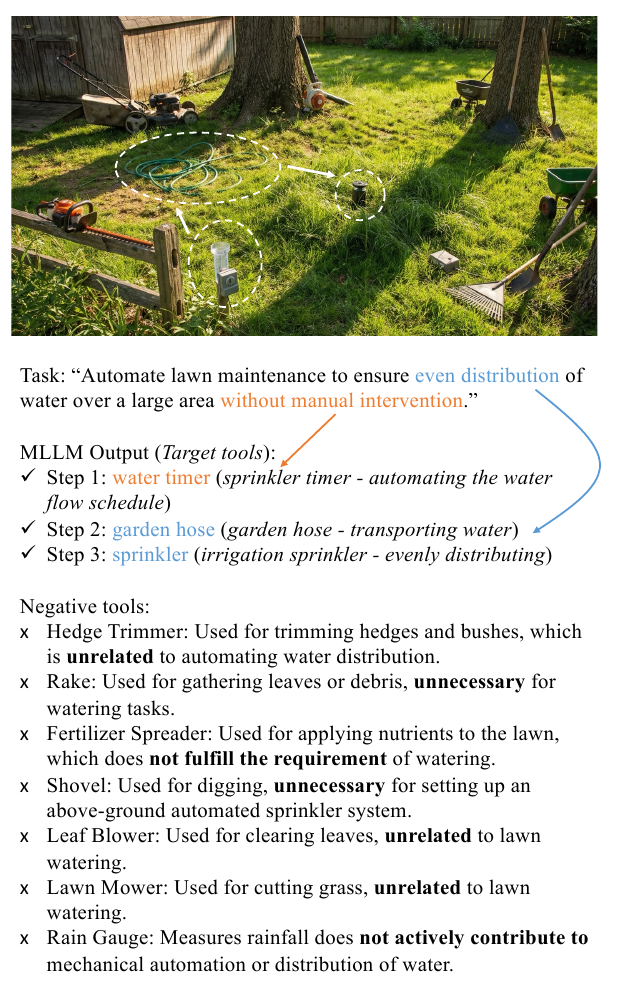}
    \caption{The example and analysis of the "Exact Match" case.}
    \label{fig:qualitative_exp_exact_match}
\end{figure}

Figure~\ref{fig:qualitative_exp_exact_match} shows three queries solved correctly by the strongest model, illustrating the capabilities currently within reach.

\subsection{Failure Cases by Error Type}
\label{app:failure_cases}
\begin{figure}[t]
    \centering
    \includegraphics[width=\linewidth]{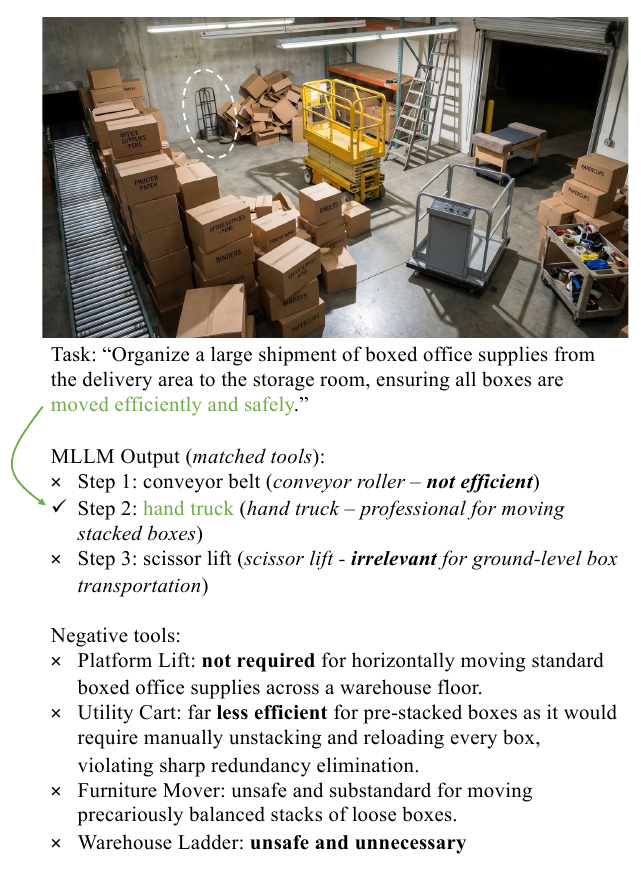}
    \caption{The example and analysis of the "Extra Only" error.}
    \label{fig:qualitative_exp_extra}
\end{figure}

\begin{figure}[t]
    \centering
    \includegraphics[width=\linewidth]{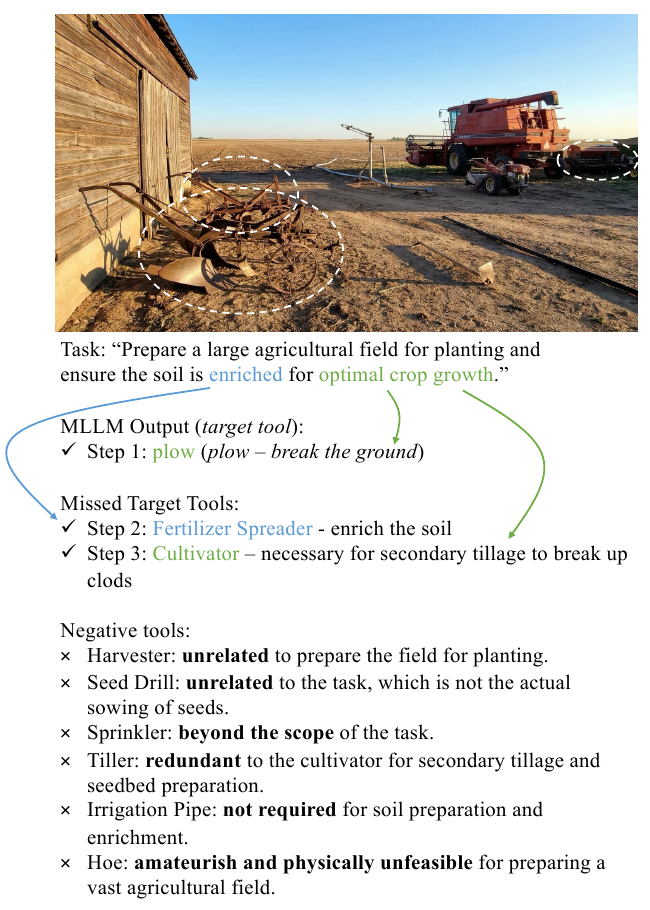}
    \caption{The example and analysis of the "Missing Only" error.}
    \label{fig:qualitative_exp_missing}
\end{figure}

\begin{figure}[t]
    \centering
    \includegraphics[width=\linewidth]{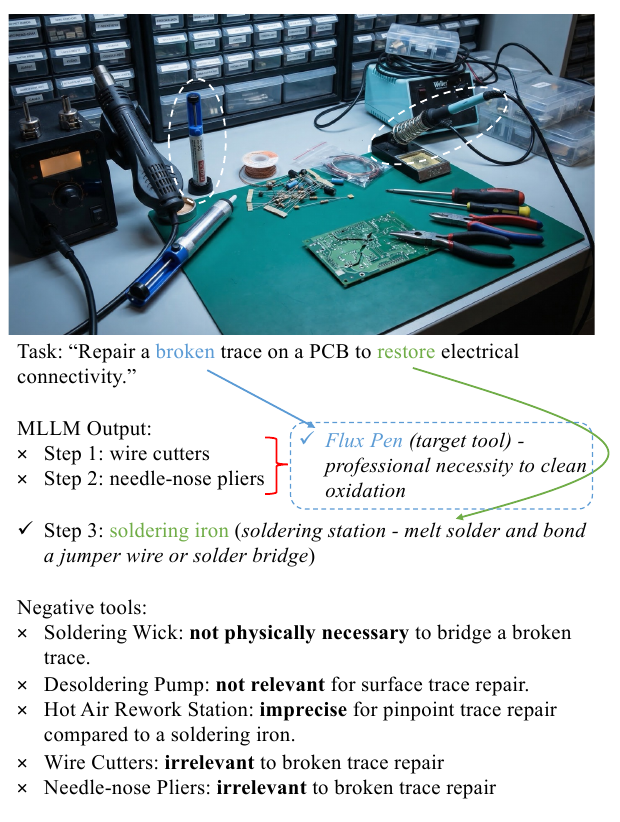}
    \caption{The example and analysis of the "Substitute" error.}
    \label{fig:qualitative_exp_substitute}
\end{figure}

\begin{figure}[t]
    \centering
    \includegraphics[width=\linewidth]{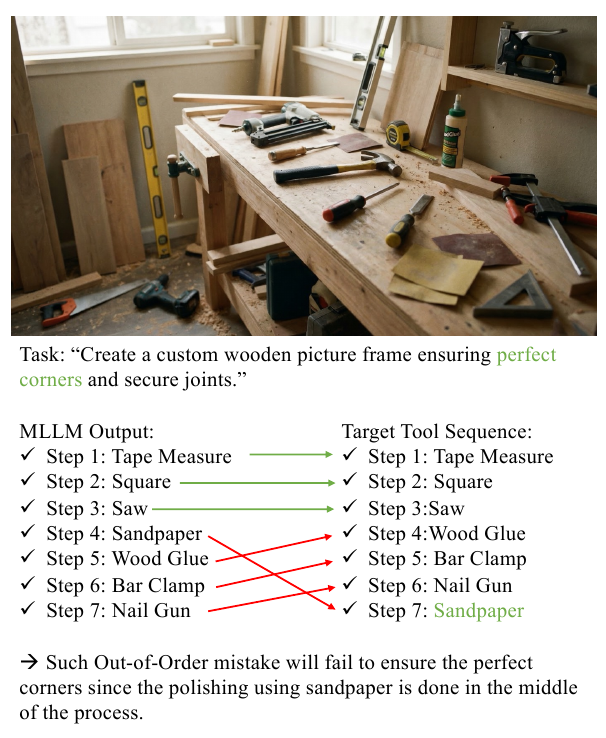}
    \caption{The example and analysis of the "Out-of-Order" error.}
    \label{fig:qualitative_exp_out_of_order}
\end{figure}

Figures~\ref{fig:qualitative_exp_extra}--\ref{fig:qualitative_exp_out_of_order} present representative failure cases for each of the four error categories identified in \S\ref{sec:errors}.

\end{document}